%% file: main.tex
\newif\ifPrePrint
\PrePrintfalse

\documentclass[letterpaper, 10 pt, conference]{ieeeconf}
\IEEEoverridecommandlockouts                              

\def\baselinestretch{0.95}

\usepackage[switch]{lineno}
\usepackage[noadjust]{cite}
\usepackage{graphics} 
\usepackage{xcolor}
\usepackage{epsfig} 
\usepackage{epstopdf}
\usepackage{amsmath} 
\usepackage{amssymb}  
\usepackage{stmaryrd}
\usepackage{transparent}
\usepackage{import}
\usepackage{bm}
\usepackage{siunitx}
\let\labelindent\relax
\usepackage[inline]{enumitem}
\usepackage[font=footnotesize,labelfont=bf]{caption}
\usepackage[export]{adjustbox}
\usepackage{subfig}
\usepackage{url}
\usepackage{hyperref}
\usepackage{bbm}
\usepackage{tikz}
\usepackage{siunitx}

\usepackage{amsthm}

\newtheorem*{obj*}{Objective}

\newtheorem*{prob*}{Problem}

\newtheorem*{assumption*}{Assumption}

\theoremstyle{remark}

\usetikzlibrary{shapes,arrows,positioning,calc}
\input{macros}

\newif\ifold
\oldfalse

\newif\ifremark
\remarkfalse

\newif\ifskeleton
\skeletonfalse

\newif\ifrevision
\revisionfalse
\ifrevision

\else

\fi

\ifPrePrint
\makeatletter
\def\ps@titlepagestyle{
	\def\@oddfoot{}\def\@evenfoot{}
	\def\@oddhead{\textcolor{red}{\sf\footnotesize Preprint version, final version at http://ieeexplore.ieee.org/ \hfill IEEE International Conference on Robotics and Automation 2021}}
	\def\@evenhead{\textcolor{red}{\sf\footnotesize  Preprint version, final version at http://ieeexplore.ieee.org/  \hfill IEEE International Conference on Robotics and Automation 2021}}%
}%
\makeatother
\makeatletter
\def\ps@headings{
	\def\@oddfoot{\textcolor{red}{\sf\footnotesize  Preprint version, final version at http://ieeexplore.ieee.org/ \qquad\qquad\quad \thepage \;\;~\hfill ~\hfill IEEE International Conference on Robotics and Automation 2021}}\def\@evenfoot{\hfill\thepage\hfill}
	\def\@oddhead{}\def\@evenhead{}%
}%
\makeatother
\fi	

\begin{document}
\pagenumbering{gobble}

%

\title{\LARGE \bf Proprioceptive measurement based variable impedance learning interaction control for aerial manipulation}

\title{\LARGE \bf 
Learning Variable Impedance Control for Aerial Sliding on Uneven Heterogeneous Surfaces \reviewchanges{by} Proprioceptive and Tactile Sensing}
\author{Weixuan Zhang, Lionel Ott,  Marco Tognon,  and Roland Siegwart
\thanks{This work was supported by the NCCR Robotics, NCCR Digital Fabrication and Armasuisse.}
\thanks{Authors are with the Autonomous Systems Lab, ETH Z\"{u}rich, Leonhardstrasse 21, 8092 Zurich, Switzerland. e-mail:
        {\tt wzhang@mavt.ethz.ch}}
}

\maketitle


\begin{abstract}
The recent development of novel aerial vehicles capable of physically interacting with the environment leads to new applications such as contact-based inspection.
These tasks require the robotic system to exchange forces with partially-known environments, which \reviewchanges{may} contain uncertainties including unknown spatially-varying friction properties and discontinuous variations of the surface geometry.
Finding a control strategy that is robust against these environmental uncertainties remains \reviewchanges{an open challenge}. 
This paper presents a learning-based adaptive control strategy for aerial sliding tasks.
In particular, the gains of a standard impedance controller are adjusted in real-time by a policy based on the current control signals, proprioceptive measurements, and tactile sensing. 
This policy is trained in simulation with simplified actuator dynamics in a student-teacher learning setup.
The real-world performance of the proposed approach is verified using a tilt-arm omnidirectional flying vehicle.
The \reviewchanges{proposed controller structure combines data-driven and model-based control methods, enabling our approach to successfully transfer directly and without adaptation from simulation to the real platform. Compared to fine-tuned state of the art interaction control methods we achieve reduced tracking error and improved disturbance rejection.}
\end{abstract}

\setcounter{section}{0}

\section{Introduction}
Aerial interaction tasks such as contact-based inspections~\cite{tognon2019truly, trujillo2019novel, watson2021dry} require the flying vehicle to slide a sensor along the surface to be inspected and maintain contact. 
The state of the art proposes several solutions \reviewchanges{for the related} general \textit{push-and-slide} problem~\cite{ollero2021past}. 
Most existing approaches show interactions with simple continuous surfaces (e.g., planes, cylinders, etc), where the geometry is assumed known and friction properties are assumed to be identical everywhere (i.e., homogeneous) ~\cite{ryll20196d, tzoumanikas2020aerial, bodie2020active}. 

However, in real applications, interaction surfaces are often discontinuous and inhomogeneous, and their model knowledge is only partial due to perception inaccuracy. 
Firstly, the surface friction property cannot be directly measured and may change discontinuously if the surface consists of different materials (i.e., heterogeneous surface).
\reviewchanges{The surface geometry estimated with visual sensors contains inaccuracies on the order of 1 cm and a few degrees in orientation.} 
Secondly, from the control perspective, 
the presence of these unexpected environment features introduces discontinuities in contact forces. 
The induced discontinuous interaction wrench causes large abrupt changes in the robot dynamics 
which easily destabilizes the system.
Thus, a control strategy capable of adapting to the unknown environment and ensure a stable flight is needed.
\reviewchanges{This work investigates methods to overcome these challenges.}



\begin{figure}[t]
  \centering
  \includegraphics[width=0.7\columnwidth]{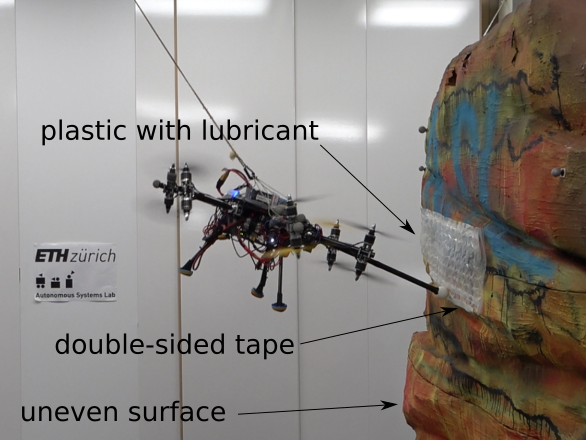}
  \caption{A tilt-arm omnidirectional flying vehicle, the Omav, is sliding along surfaces with different friction properties and geometries. The images shows the end effector is in contact with a double-sided tape while sliding on a rock like papier-m\^ach\'e.}
  \label{pic:Omav_sliding_rock}
\end{figure}


Literature on adaptive sliding on partially-known and uneven surfaces exists mainly in the manipulation community, using passivity \cite{shahriari2019power} or adaptive force control \cite{jung2004force, amanhoud2020force}, but \reviewchanges{they do not directly transfer to aerial robots}. 
\reviewchanges{Adaptive force control is often used to modify the interaction force along the direction normal to the surface according to geometry changes, while keeping the force constant along the sliding direction.} As a result, an adaptation to heterogeneous surface is not possible.
One could also employ disturbance observer-based robust control \cite{lee2020aerial}. Such an approach has the disadvantage of being slow to react, especially in the presence of noisy measurements and inaccurate process models. Consequently, the flying vehicle will struggle to handle abrupt changes in the environment.
Finally, another option is to use mechanical compliance at the flying machine's end effector~\cite{suarez2018physical}. However, this increases the mechanical complexity and cost while further decreasing the system's  payload. \reviewchanges{Furthermore, it remains limited in terms of adaptability to environmental conditions.}

An alternative approach to address the aforementioned sensing challenge is to use proprioceptive measurements and tactile sensing to react to changes in the geometry and friction properties of the environment.
This type of approach can be found in quadrupedal robots like the MIT Cheetah \cite{hyun2014high} and Anymal \cite{lee2020learning}. These robots use either control signals to infer a leg touchdown event or use IMU signals, robot states, and control commands to implicitly infer the surface properties. 
Compared to 3D visual sensors, these two sensing modalities are more direct in measuring the contact and are more suitable in sensing the discontinuous features of the environment.
\reviewchanges{However, they were never exploited in the field to aerial manipulation to face unknown environments.}

From a control perspective, impedance control \cite{hogan1984impedance} is a suitable control strategy for physical interaction tasks, as it introduces algorithmic compliance, \reviewchanges{making} the controlled system less susceptible to disturbances. This has been successfully demonstrated for sliding tasks on homogeneous surfaces with aerial vehicles \cite{ryll20196d,bodie2020active} or human-robot interaction \cite{tognon2021physical}.
Selection of the impedance gains is a trade-off between controller tracking performance and system compliance. 
\reviewchanges{Depending on the task, different impedance parameters may perform optimally \cite{walker2011demonstrating}.}
\reviewchanges{For this reason, a variable impedance controller \cite{ikeura1995variable} is an attractive solution.} 
\reviewchanges{The works in \cite{mersha2014variable} and \cite{bodie2020active} employed variable impedance control for aerial manipulation, recently formulated in terms of passivity-based control \cite{gerlagh2020energy, benzi2022}. }

\reviewchanges{A straightforward application of variable impedance control in sliding applications, is to change the stiffness gain according to the estimate of the friction coefficient obtained from a force-torque sensor. 
However, a low-pass filter for the noisy force-torque sensor measurements may introduce delay and the time-varying sensor bias may lead to incorrect estimates during flight.
Another way would be to infer an adapting gain policy directly from the sensors reading, without passing though an explicit estimation of the environment parameters.}

\reviewchanges{With recent advances in machine learning, deep reinforcement learning (RL) techniques \cite{sutton2018reinforcement} have become a popular tool to generate highly nonlinear and effective control policies using neural networks. 
Learning from simulation instead of from real-world is the preferred approach, especially concerning flying vehicles where a failure typically leads to a crash.} 
In particular, student teacher setups \cite{chen2020learning} are exploited for improved learning efficiency, since a teacher policy can use privileged information from simulation to guide the student policy.
On one hand, there exists a large quantity of works on end-to-end (i.e, from perception to actuator commands directly) reinforcement learning from simulation (see \cite{polydoros2017survey} and references therein). While being powerful, they are generally not efficient and require a high-fidelity simulation of the environment.
On the other hand, methods combining reinforcement learning with variable impedance control have been proposed. Examples include manipulator control for light switch turning \cite{{buchli2011learning}}, peg-in-hole tasks \cite{beltran2020variable,beltran2020learning}, human-robot collaboration \cite{roveda2020model}, and hopper jumping \cite{bogdanovic2020learning}. 
They keep the impedance controller structure in the learning framework improving data-efficiency and requiring less fidelity in the simulation.






\reviewchanges{Considering the problem of aerial physical interaction with unknown surfaces, taking inspiration from different robotic domains, this paper presents a novel control method that combines the benefits of proprioceptive and tactile sensing, variable impedance control and reinforcement learning.} 
Contrary to the previously mentioned works, \reviewchanges{instead of estimating the friction coefficients or surface geometry,} we directly learn a mapping from the proprioceptive and tactile signals to the impedance control gains.
\reviewchanges{When the environmental property changes, this neural network policy rejects these disturbances, preserving contact and keeping the orientation of the flying vehicle's end effector constant. In particular, the policy changes the impedance control parameters according to tracking errors, IMU sensor measurements, and wrench sensing.}
A key hypothesis here is that there exists an underlying mapping between these signals and the control gains.
The training of this policy is conducted entirely in simulation \reviewchanges{exploiting the simplified closed-loop dynamics, in a student-teacher setup.
This solution allows a direct transfer of the learned policy from simulation to an omnidirectional aerial vehicle (Omav), significantly improving its robustness during interaction (Omav as shown in Fig.~\ref{pic:Omav_sliding_rock} and described in \cite{bodie2019omnidirectional}).}

\reviewchanges{Our contributions can be summarized as follows:
\begin{itemize}
  \item An aerial sliding control strategy that adapts its gain to reject disturbances present in the interacting environment, including discontinuous changes in surface friction and geometry;
  \item Insights on how to address sim-to-real transfer by including a closed-loop controller to suppress model uncertainty and learning from simplified actuator dynamics for aerial interaction tasks.
\end{itemize}
The above contributions have been validated in the experiments using an Omav with a rigid single-body end effector for the task of sliding across challenging surfaces including a step and discontinuous surface friction changes. }





\section{Preliminaries}

In this section we provide a brief overview of the models used to represent robot dynamics and environment interactions before describing the basics of impedance control.

\subsection{Robot dynamics}
\begin{figure}[t]
  \centering
  \includegraphics[trim=0 0 0 0, clip,width=\columnwidth]{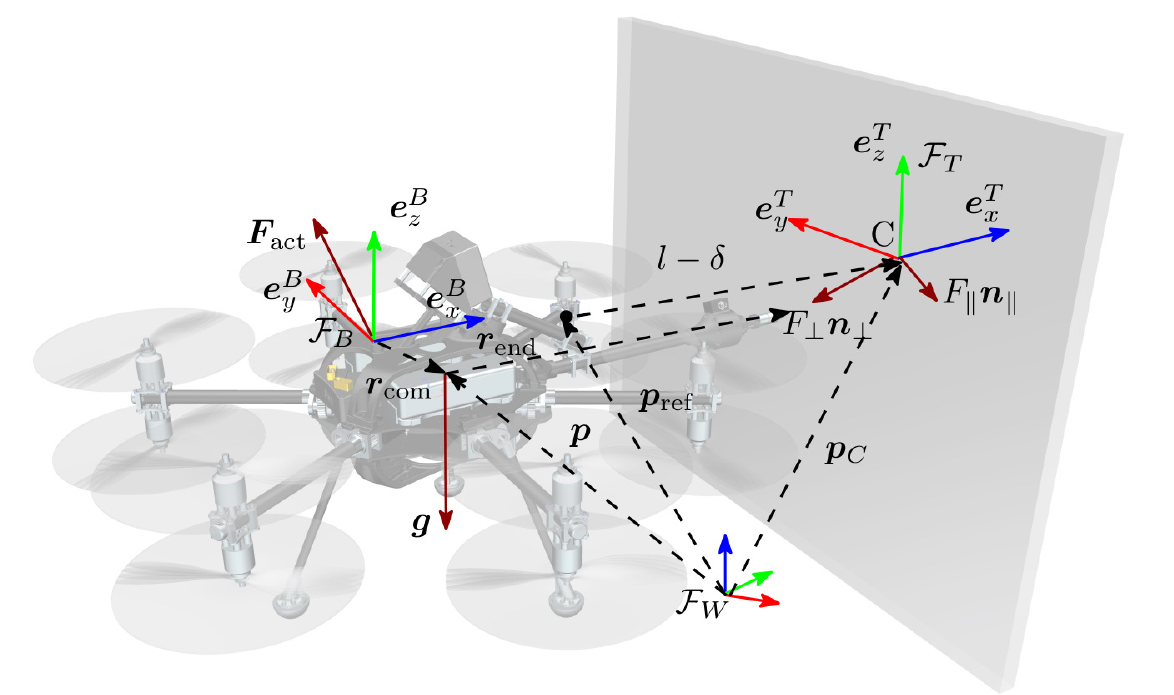}
  \caption{Omav in interaction, showing some of the symbols and quantities required to model the system}
  \label{pic:Omav_notation}
\end{figure}

It is assumed that the flying vehicle has a single-body end effector rigidly attached to its body (see Fig.~\ref{pic:Omav_notation}).
The robot is modeled as a single rigid body and its dynamics is expressed \reviewchanges{using Newton-Euler method} in free flight and interaction are given by the following equation
\begin{equation} \label{eq:lagrangian}
\bm{M} \reviewchanges{\generalizedAcc} + \bm{C} \reviewchanges{\generalizedVel} + \bm{g} = \wrench{\act} + \wrench{\dist},
\end{equation}
\noindent where $\bm{M} \in \mathbb{R}^{6 \times 6}$ is the symmetric positive definite inertia matrix and $\bm{C} \in \mathbb{R}^{6 \times 6}$ contains the centrifugal and Coriolis terms, and $\bm{g}\in \mathbb{R}^6$ is the gravity.
\reviewchanges{The generalized velocity $\generalizedVel \in \mathbb{R}^6$ represents the center of mass velocity and body rates of the system.}
The generalized acceleration vector are denoted as $\generalizedAcc$.
The terms $\wrench{\act}$ and $\wrench{\dist} \in \mathbb{R}^{6}$ are both stacked force and torque vectors acting on the system generated  by rotor actuation and disturbance sources (e.g., contact or wind disturbances), respectively.

\subsection{Interaction with the environment}
When the robot is sliding along the surface with its end effector,
the force disturbance $\mvec{f}_\dist$ has three sources: \begin{enumerate*}
\item environmental aerodynamic effects, e.g. ground effects, wall effects, and wind gusts, 
\item actuation modeling errors \cite{zhang2020learning}, and 
\item the contact force $\mvec{f}_\contact$.
\end{enumerate*}
Wall effects and wind gusts are assumed to be negligible for indoor flights as they account for less than one percent of prediction error according to \cite{garofano2021aerodynamic}.
The disturbance caused by modeling errors is an order of magnitude smaller than that of the contact force and thus also assumed to be negligible.

During interaction flights, the robot \reviewchanges{is assumed to have} a single contact point with the \reviewchanges{uneven} surface at $C$ (Fig.~\ref{pic:Omav_notation}). \reviewchanges{A local contact frame $\mathcal{F}_T$ is attached to the contact point  such that its $x$-axis is normal to the tangent plane at $C$.}
The contact force acting on the end effector expressed in the body frame $\mathcal{F}_B$ is modeled as follows:
\begin{equation}
  \mvec{f}_\dist^\base = \mvec{f}_\contact^\base = \rot{\base}{\tool} (F_\perp \vec{n}^\tool_\perp +  F_\parallel \vec{n}^\tool_\parallel),
\end{equation}
where $F_\perp$ is the scalar normal force and $F_\parallel$ is the scalar friction force.
The coordinate transformation matrix $\rot{\base}{\tool}$ transforms a vector from $\mathcal{F}_T$ to $\mathcal{F}_B$.
The unit normal vector $\vec{n}^\tool_\perp$ is perpendicular to \reviewchanges{the tangent plane at $C$}
and $\vec{n}^\tool_\parallel$ is the sliding force direction \reviewchanges{parallel to the tangent plane at $C$}.
The relative orientation $\rot{\base}{\tool}$ is assumed to be partially known due to imperfect map.
In addition, the end effector and the contact force creates a torque around the vehicle's center of mass with the lever arm length denoted as $l$.

When the end effector is sliding with nonzero velocity on the surface, a Coulomb friction model is assumed, i.e.:
\begin{equation}
  F_\parallel = \mu(\vec{p}_C) F_\perp,
\end{equation}
where $\mu(\vec{p}_C)$ is the friction coefficient that can vary spatially across the surface, depending the position of contact $\vec{p}_C$.

\subsection{Impedance controller with constant gains}
An impedance controller with constant control gains is a common approach used for aerial sliding tasks \cite{bodie2019omnidirectional} and used in this paper as the baseline approach.

Given a desired sliding path on the surface, a reference pose trajectory is designed based on the \priorMap. This reference trajectory consists of a desired center of mass position which results in a end effector position that is always behind the sliding surface by a constant distance $\delta \in \mathbb{R}$ (Fig.~\ref{pic:Omav_notation}), also denoted as the penetration level. 
For the attitude part of the trajectory, the vector along the tool arm should align with the contact frame $x$-axis.

Given this desired reference trajectory, 
an impedance controller with constant control gains has the following form:
\begin{equation} \label{eq:impedance1}
    \begin{split}
    \wrench{\act} &= \bm{C}\generalizedVel + \bm{g} + (\bm{M}\desInertia^{-1} - \mathbb{I}_6)\wrench{\ext} \\
    & - \bm{M}\desInertia^{-1}(
    - \inertia \generalizedAccRef + 
    \desDamping \generalizedVelError+ \desStiffness \generalizedPosError),
    \end{split}
\end{equation}
with $\generalizedPosError \in \mathbb{R}^6$, containing the position and attitude tracking error as shown in \cite{bodie2019omnidirectional}. 
$\desInertia, \desDamping, \desStiffness \in \mathbb{R}^{6 \times 6}$ are the desired inertia, damping, and stiffness matrices, respectively.

Plugging \eqref{eq:impedance1} into \eqref{eq:lagrangian} results in
\begin{equation} \label{eq:desired_dynamics_baseline}
\desInertia \generalizedAccError + \desDamping \generalizedVelError + \desStiffness \generalizedPosError = \wrench{\ext}.
\end{equation}
This implies that an impedance controller shapes the closed-loop system as a second-order system.
Note that there will be inevitable pose error due to the contact wrench acting on the flying vehicle.
The wrench command $\wrench{\act}$ is then allocated through a chosen mapping and a saturation function to individual actuator commands (for more details see \cite{bodie2019omnidirectional}).




\section{Methodology}


\subsection{Problem statement} \label{subsec:problem_statement}

The goal is to enable a flying vehicle to accurately follow a trajectory \reviewchanges{planned based on an imperfect map} while \reviewchanges{remaining stable and} staying in contact with a surface which has unknown discontinuities in geometry and unknown friction properties $\mu(\vec{p}_C)$. \reviewchanges{Given a task-space reference trajectory, we assume the robot has access to contact wrench measurements $\wrench{\meas}$ via a force torque sensor, and it is controlled by an impedance controller with a gain-adjusting policy $\pi_\theta$, which is parametrized by $\theta$.} To achieve the above goal, we propose a strategy to \reviewchanges{find a deterministic policy $\pi_\theta$ that adjusts the controller's gains to fulfill the following criteria:
\begin{enumerate*}
    \item minimize the tracking error $\norm{\generalizedVelError}^2 + \norm{\generalizedPosError}^2$ where $\norm{\cdot}$ denotes the Euclidean norm,;
    \item ensure $\mvec{f}_\contact^\base > 0$;
    \item ensure the platform stability.
\end{enumerate*}
}



\subsection{Simulation using simplified dynamics} \label{subsec:simplified_sim}
To allow for efficient evaluation and training of the policy $\pi_\theta$, a simplified dynamics simulation is used.
The flying vehicle is simulated as a single rigid body and the simulation of the individual actuator dynamics are approximated collectively as a single process.
A saturation function on the wrench command is implemented, the output of which is delayed and set as external force and torque directly acting on the robot.
Both the saturation threshold and the system delay are a conservative estimate of the empirically obtained actuation limits.
This ensures that the actuator limits are well respected and the closed-loop system behaves like a delayed second-order system as designed.
The inertia and mass are obtained from CAD.
Although the actuator dynamics are simplified, special attention is paid to identify the correct center of mass position and the relative position of the end effector in the body frame ($\bm{r}_\mathrm{com}$ and $\bm{r}_\mathrm{end}$ in Fig.~\ref{pic:Omav_notation}). 
They together determine the induced torque disturbance from a given contact force, which is essential for the simulation to learn the correct disturbance rejection strategy.


To simulate the interaction environment, surfaces that have different friction coefficients\footnote{The same material was used for the end effector throughout this paper. Thus, for the sake of brevity, we only talk about surface friction coefficients when it would be more accurate to talk about friction pairs between the end effector and the surface.} are generated and concatenated together.
Therefore, when the robot's end effector slides across the border between two surfaces with different friction properties, it experiences discontinuous changes in interaction forces.
Furthermore, each surface can have a different height that leads to an uneven surface as a whole.


\subsection{Variable impedance learning controller}


\begin{figure} 
    \center
     \includegraphics[width=\columnwidth]{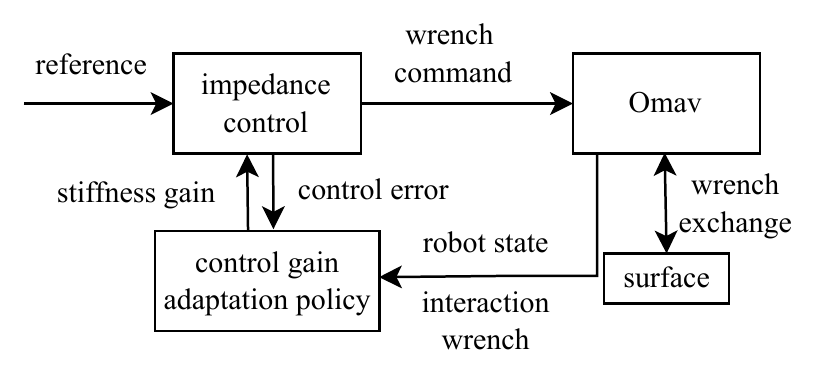}
     \caption{Variable impedance learning controller. \reviewchanges{It augments the control strategy depicted in \cite{bodie2019omnidirectional} by adding a control gain adaptation policy}, which takes as input the state, control error and the wrench measurements from the wrench sensor and map them to impedance gain.}
    \label{pic:control_tikz}
 \end{figure} 
 
The proposed approach \reviewchanges{adds a control gain adaptation policy to the standard impedance controller \eqref{eq:impedance1}. The policy} adapts the impedance controller gains based on the proprioceptive measurements and the tactile feedback via the adaptive unit described in Fig.~\ref{pic:control_tikz}. 
In particular, the desired stiffness $\desStiffnessAdaptive \in \R{6 \times 6}$ is a function of the $m$-dimensional measurements $\feature \in \R{m}$:
\begin{equation} \label{eq:adaptiveStiffness}
  \desStiffnessAdaptive = \desStiffnessMin + (\desStiffnessMax - \desStiffnessMin) \operatorname{diag}\{\pi(\feature)\}
\end{equation}
where $m$ is the dimension of the measurements, $\desStiffnessMin$ and $\desStiffnessMax$ are diagonal positive semidefinite matrices, and $\pi: \R{m} \rightarrow \R{6}$ where $\zeroVector < \pi(\cdot) < \oneVector$ with vector $\zeroVector$ and $\oneVector$ of dimension 6.
These constraints on the mapped value make sure the adaptive gains have lower and upper bounds.
The lower bound $\desStiffnessMin$ ensures a minimum tracking performance while the upper bound $\desStiffnessMax$ prevents the system from instabilities caused by actuator saturation and system delay. 
These limits are both derived empirically through experiments.
\reviewchanges{For brevity, in the following we use $\pi$ instead of $\pi_\theta$.}

To obtain a damped second-order system, we impose a fixed relationship between $\desDamping$ and $\desStiffnessAdaptive$. The desired damping $\desDamping$ is varied with the square root of the diagonal components of $\desStiffnessAdaptive$:
\begin{equation} \label{eq:damping}
\desDamping = 2\zeta \sqrt{\desStiffnessAdaptive},
\end{equation} 
where $\zeta$ is a damping ratio. 
While we assume that the desired stiffness and damping can be well tracked, 
the desired inertia is in practice challenging to track as it requires \reviewchanges{an accurate actuation control \cite{ryll20196d}.} 
We therefore set $\desInertia$ equal to $\inertia$ and the adaptation of $\desInertia$ is deferred to future works. 

With $\inertia = \desInertia$ and \eqref{eq:damping} inserted into \eqref{eq:impedance1}, the following adaptive controller command can be obtained,
\begin{equation} \label{eq:impedance2}
    \wrench{\act} = \inertia \generalizedAccRef - 2\zeta \sqrt{\desStiffnessAdaptive} \generalizedVelError - \desStiffnessAdaptive \generalizedPosError + \bm{C}\generalizedVel + \bm{g}.
\end{equation}
With \eqref{eq:impedance2} plugged into \eqref{eq:lagrangian}, the closed-loop error dynamics are shaped as a second-order system,
\begin{equation} \label{eq:desired_dynamics}
\desInertia \generalizedAccError + 2\zeta \sqrt{\desStiffnessAdaptive} \generalizedVelError + \desStiffnessAdaptive \generalizedPosError = \wrench{\ext}.
\end{equation}

\reviewchanges{
Note that changing the stiffness $\desStiffnessAdaptive$ affects the interaction wrench $\wrench{\ext}$. To see this, consider \eqref{eq:desired_dynamics} at steady state $\generalizedAccError = \generalizedVelError = \bf{0}$ and projected along the surface normal direction. We obtain}
\begin{equation}\label{eq:steady_state}
  k_\perp(\feature) \delta = F_\perp,
\end{equation}
where $k_\perp(\feature)$ is the position stiffness gain in the surface normal direction.
Assuming the end effector is in contact with the surface, which results in a constant position tracking error $\delta$, $F_\perp$ is thus proportional to $k_\perp(\feature)$.  
This is the intuition of why we adapt the stiffness gain to counteract surface disturbances.






\reviewchanges{
\subsection{Reinforcement learning of gain adaptation policy }
We assume that the mapping $\pi$ can be modeled as a discrete-time continuous Markov decision process (MDP).
An MDP is defined by a state space $\mdpState$, an action space $\mdpActionSpace$, a scalar reward function $\mdpRewardSpace$, and the transition probability $\transitionProb$ that dictates the stochastic system dynamics. 
A learning agent selects an action $\action$ from its policy $\pi$ and receives a reward $\reward$.
The objective of the RL framework is to find an optimal policy $\pi^*$ that maximizes the discounted sum of rewards over an infinite time horizon:
\begin{equation}
    \pi^* = \arg \max_\pi \mathbb{E}_{\tau(\pi)} \sum_{t=0}^\infty \gamma^t \reward[t] 
\end{equation}
where $\gamma \in (0, 1)$ is the discount factor,  and $\tau(\pi)$ is the trajectory distribution under policy $\pi$, with $t$ denoting the discretized time indices.
The reward $\reward[t]$ at $k$ is
\begin{equation}\label{eq:reward}
    \begin{split}
    &  \reward[t] = - \weight_{\vec{e}_{R}}  \norm{\vec{e}_{R}[t]}^2  - l_p \norm{\bm{e}_p[t]}^2  - \reviewchanges{\weight_{\eDist}} \norm{\eDist[t]}^2   \\ & - \weight_{\vec{\omega}} \norm{ \vec{\omega}[t]}^2  -  \reviewchanges{\weight_{\action}} \norm{\frac{\action[t]}{\norm{\action[t]}} - \frac{\action[t-1]}{\norm{\action[t-1]}}}^2 
    \end{split}
\end{equation}
 $\weight_{\star}$ with subscripts $\star$ denotes the corresponding weight chosen such that all individual reward terms are at the same order of magnitude. 
$\vec{e}_{R}[t]$ and $\vec{e}_{p}[t]$ denotes the attitude and translational tracking error.
$\eDist[t]$ denotes the scalar distance along the $x$-axis of the local contact frame between the end effector and the surface.
\reviewchanges{The purpose of this term is to make sure that the policy keeps the end-effector in contact with the surface.}
$\vec{\omega}[t]$ denotes the angular velocity. \reviewchanges{We penalize large angular velocities to avoid instabilities.}
The action $\action[t]$ is the output of the policy $\pi(\feature)$ at time $t$. The associated term is to make sure that the control inputs are smooth.
The loss components are designed to reduce the tracking error while keeping contact with the surface without causing discontinuities in the actions and thus the actuator commands.
These terms are in line with the problem statement in Sec.~\ref{subsec:problem_statement}.  
Note that since the simulation of each individual actuator dynamics is omitted. The energy consumption of the flying vehicle can only be indirectly inferred and is therefore not included in the reward function.
}

\reviewchanges{
The policy $\pi$ is a fully connected neural network with three hidden layers of 32 units, its activation functions being leaky ReLu, and its last layer being a Sigmoid layer which guarantees to map to a bounded interval. For training we use the off-the-shelf RL algorithm proximal policy optimization (PPO) \cite{schulman2017proximal}, a policy gradient algorithm that has been demonstrated to work for variable impedance control in contact tasks with a manipulator \cite{martin2019variable}.
}

\subsection{\reviewchanges{Learning from simulaiton}}
To efficiently learn an optimal policy that determines the adaptive stiffness $\desStiffnessAdaptive$ (see~\eqref{eq:adaptiveStiffness}), a student teacher learning approach \cite{chen2020learning} is deployed.

\begin{figure} 
    \center
     \input{images/learn_from_simulation}
     \caption{\reviewchanges{This block diagram shows the main steps in the student-teacher learning of the control gain adaptation policy.}}
    \label{pic:student_teacher_learning}
 \end{figure}
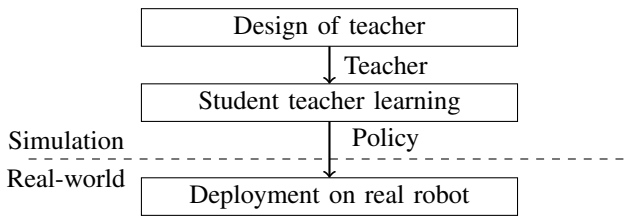

Fig.~\ref{pic:student_teacher_learning} provides an overview of this approach:
\reviewchanges{Firstly, we design a teacher with access to privileged (ground-truth) information to dynamically select the desired stiffness in the variable impedance controller. 
Then a policy is learned to emulate the teacher and may be further improved using RL.
The policy can be directly transferred to real-world without any additional sim-to-real adaptation..
}
\reviewchanges{
The intuition behind the student teacher learning is that the teacher has access to the privileged information is much easier to design or train in an RL setting.
We can also embed empirical tuning experience or other adaptive variable impedance strategies into the teacher.
This is helpful for challenging tasks, as we find out empirically a direct reinforcement learning always lead to instabilities of the flying vehicle and prevents successful learning.
}


\subsubsection{Teacher design} \label{subsubsec:teacher_policy}
The teacher $\teacherPolicy$ serves as a guidance policy for the student policy. It makes use of the ground truth information $\privilegedInfo$ (friction coefficients and local height map) from the simulation, to which the student policy does not have access in real deployment. 

In this work, \reviewchanges{we employ either a simple handcrafted policy $\pi_\mathrm{t}$ or a neural network learned from simulation using reinforcement learning} \reviewchanges{as the teacher}.
\reviewchanges{For handcrafted policy, the design intuition of the handcrafted policy is mainly based on the fact that with a larger friction coefficient, the angular stiffness gain should be increased to counteract the increased torque disturbance. At the same time, the translational stiffness gain should be decreased to reduce normal force $F_\perp$ (see~\eqref{eq:steady_state}) and the torque disturbance. 
For reinforcement learning of the teacher, friction coefficient at contact point and the surface normal vector in the neighbourhood are used in addition to $\feature$.
They directly provide information about the interacting environment: the surface unevenness and friction property.
}

\subsubsection{Student learning} \label{subsubsec:student_policy}
\reviewchanges{The control gain adaptation policy $\initialStudentPolicy(\feature)$ is bootstrapped via supervised learning with the following loss function
\begin{equation} \label{eq:supervisedLoss}
    \initialStudentPolicy = \arg \min_\pi \norm{\teacherPolicy(\privilegedInfo) - \pi(\feature))}^2.
\end{equation}
where the feature $\feature$ contains control signal, state estimate, interaction wrench measurements and the IMU signals. Here, the control signal, the IMU measurements and the state estimate are the propprioceptive sensing and interaction wrench measurements are tactile sensing. They indirectly provide the information about the interaction between the environment and the robot.}

Training data is collected by rolling out the simulation using the teacher.
For each rollout, the robot first approaches the surface, gets into contact and starts sliding following the desired trajectory.
The policy $\initialStudentPolicy(\feature)$ can be further refined using RL. 





\subsubsection{Data processing}
Noisy observations (especially the interaction wrench measurements) are first-order low-pass filtered before they are input to the student policy. 
To have a smooth change in the stiffness gain $\desStiffnessAdaptive$, the output of the student policy is also low-pass filtered before they are used to compute $\desStiffnessAdaptive$.

\subsection{Remark}
\paragraph{Stability}
\reviewchanges{We remark that the stability using RL may be ensured using the concept of passivity \cite{spyrakos2019passivity}, or Lyapunov methods \cite{khader2020stability}.
This is left as future work.
Instead we discuss here practical measures to face possible causes of the instability. Those are mainly twofold: actuator saturation due to the rapid changes in the control gains or high gains and low gains which is incapable of stabilizing this open-loop unstable system.
Thus we implemented the following strategies:
\begin{enumerate*}
  \item Lower and upper bounds on the stiffness gain as shown in \eqref{eq:adaptiveStiffness};
  \item Slew rate limit on the gains are empirically determined;
  \item The output of the policy $\optimalStudentPolicy(\feature)$ is filtered for a smooth control signal.
\end{enumerate*}
}

\paragraph{Sim-to-real transfer}
The sim-to-real gap, i.e., the mismatch between simulation and reality, is a challenging problem when learning from simulation, and can limit the transferability of the learned policy.
Frequent causes of sim-to-real gaps are inaccurate modeling of the actuator dynamics and delays in the system \cite{hwangbo2019learning}. 
This is especially a problem for end-to-end learning approaches. 
Without feedback controller in the loop the learning procedure relies on the accuracy of the open-loop dynamics simulation and the sim-to-real gap can diverge exponentially.
However, given a well-designed feedback controller (e.g., the one one presented in Fig.~\ref{pic:variable_impedance_control}), which shapes the system to a desired second-order system \eqref{eq:desired_dynamics} and suppresses model uncertainty, 
the gap between the reality and the simulation is kept small.
This is particularly advantageous for a complicated system like the Omav, where a large amount of training data is required for an accurate model learning of the whole body dynamics (18 actuators and 6 degrees of freedom). As a comparison, a million samples are required for the modeling of a single one degree of freedom actuator of quadrapedal walking robot \cite{hwangbo2019learning}. 
What is worse, if the Omav crashes or if its configuration changes, training data needs to be recollected again for an accurate model learning.
Our approach does not require such a meticulous effort.


\section{Experiments}
\subsection{Experimental setup}
\reviewchanges{The experimental set-up and the platform are shown in Fig.~\ref{pic:Omav_sliding_rock}.} 
The experiments are carried out at the indoor aerial robotic testbed of the Autonomous System Lab, ETH Zurich. 
The Omav weighs $4.5 \,\rm kg$ and is equipped with an NUC i7 computer and a PixHawk flight controller. For a more complete description of this platform see \cite{bodie2019omnidirectional}. 
A pole end effector is rigidly attached to the vehicle and points outwards for a sufficient margin between the propellers and the surface.
A ping pong ball is attached at the pole tip to ensure a single contact during sliding movements. 
A force torque sensor from Rokubi, mounted to the end effector, measures the interaction wrench. 
\reviewchanges{There are three interaction environments for real experiments: sand paper on a flat white board to investigate interaction with heterogeneous surface, a step of \SI{2}{\cm} on a white board (Fig.~\ref{pic:Omav_sliding}) to investigate interaction with discontinuous surface geometry, and finally, a rock-like structure (Fig.~\ref{pic:Omav_sliding_rock}) which combines both traits to form a challenging environment typically seen in real applications.
A motion capture system provides pose estimates for both the robot and the whiteboard/rock at $100\,\rm Hz$.
The robot is unaware of their surface properties and the local unevenness on the surface. 
The task trajectory is to follow a straight line trajectory parallel to the gravity vector with zero pitch and roll while sliding on the vertical surface. 
The penetration level $\delta$ is set to \SI{0.07}{\meter}, which leaves sufficient margin to ensure contact under the state estimate uncertainty.}

For the simulation we use RaiSim \cite{raisim}, a cross-platform multi-body physics engine for robotics. 
During training in the simulation, for each rollout, the robot approaches the surface and starts sliding along the surface for 15 seconds, which emulates a task trajectory from the real-world experiments.
During each rollout, the Omav reaches a speed of \SI{0.2}{\meter\per\second} and slides across surfaces with different friction coefficients.  
For each interaction environment, we set up a different interaction environment in the simulation and learn a different policy to treat each problem separately. 
Find a single policy that tackles different environments can be studied in the context of continual learning and is planned for future work.


To train the teacher or student policy using RL, a hundred simulation instances are spawned with slightly perturbed vehicle and environment properties.
In each epoch, these instances are simultaneously simulated to obtain a batch of rewards for stochastic gradient policy optimization. 
The policy was trained on a single NVIDIA 3060Ti GPU. \reviewchanges{which takes from 2 to 8 hours depending on the interacting environment.}

Since there is only a straight line to follow, the dimension of the measurement vector $\feature$ is reduced to six: filtered pitch velocity, pitch error, position control error in the sliding direction, linear velocity, friction force filtered, normal force filtered. 
The output to the action space are the linear gain in the surface normal direction and the angular gain in the pitch direction, i.e. the axis of torque disturbance.
The rest of the controller gains is kept constant.

\subsection{Sim-to-real transfer}
\begin{figure}[t]
\centering
\includegraphics[width=0.8\columnwidth]{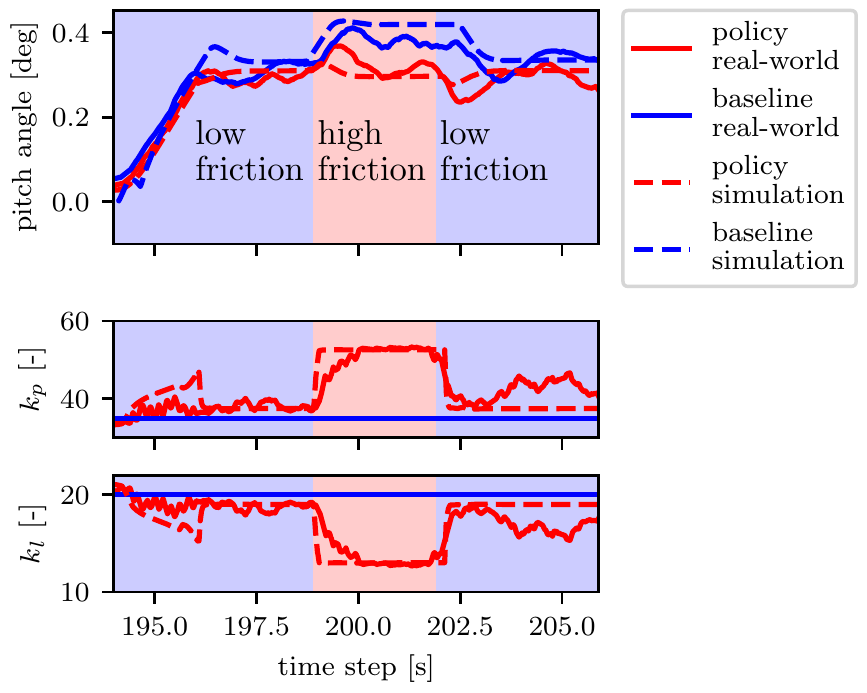}
\caption{Comparison between baseline and our approach while the robot approaches and slides across two surfaces with different friction coefficients, both in simulation (dashed line) and real-world (solid line). 
$k_a$ and $k_l$ denote the angular stiffness gain and translation stiffness gain, respectively.
}
\label{pic:simToReal}
\end{figure}

Fig.~\ref{pic:simToReal} demonstrates the policy transfer from simulation to reality.
It compares four experiments completing the same sliding task with different controllers (baseline or our approach) and different set-ups (simulation or real experiment).
The plots are aligned using the measured force impact when the Omav enters from free flight into contact with the whiteboard. 

For evaluation, the Omav must slide across three concatenated surfaces: the first is the whiteboard with low friction, then a sand paper with high friction, and finally the whiteboard again.
The friction coefficients are empirically estimated through force torque sensor measurements, with which three surfaces are generated in the simulation that replicate the experimental evaluation set-up.
The approach in evaluation is a policy $\initialStudentPolicy$ defined in the \eqref{eq:supervisedLoss} (without further refinement using RL) and a baseline impedance controller with constant control gains.
The policy is trained in a simulation environment with six different surface friction coefficients \reviewchanges{(0.05, 0.15, 0.25, 0.45, 0.55, 0.62)} that covers a range of friction coefficients. During collection of data using the teacher, each surface is randomly assigned one of the six friction coefficients to robustify the learned policy.

Fig.~\ref{pic:simToReal} demonstrates that the learned adaptive control strategy can be successfully transferred from simulation to the real-world. Given the same controller (either baseline or our approach), the simulation and the real-world experiments result in a close similarity of the respective robot state (the pitch angle) and controller gains (the angular stiffness gain $k_a$ and translational stiffness gain $k_l$).
The pitch angle is shown since it has the most obvious correlation with the surface friction coefficient given the baseline controller.


\subsection{Sliding across a heterogeneous flat surface}\label{subsec:diffCoeff}
Fig.~\ref{pic:simToReal} also showcases the performance of the regressed student policy on heterogenous surfaces.
In this subsection, we only evaluate the real experimental data.
While sliding on the surface, the contact force unavoidably leads to a tilted angle of the Omav.
With the baseline controller, the vehicle tilts on average \SI{4.9}{\degree} after encountering a high friction surface (\reviewchanges{from \SI{199}{\second} to \SI{202}{\second}}), whereas with our approach, the tilt of the vehicle only increases to about  \SI{1.3}{\degree} on average. 
This shows that the Omav is able to keep a almost constant tilt angle when sliding across different surfaces, thus improve the attitude tracking performance and reducing the chance of a crash at the transition of difference surfaces.

\subsection{Sliding over a surface with discontinuous geometry}


\begin{figure}[t]
\centering
\subfloat{\includegraphics[width=0.5\columnwidth,valign=c]{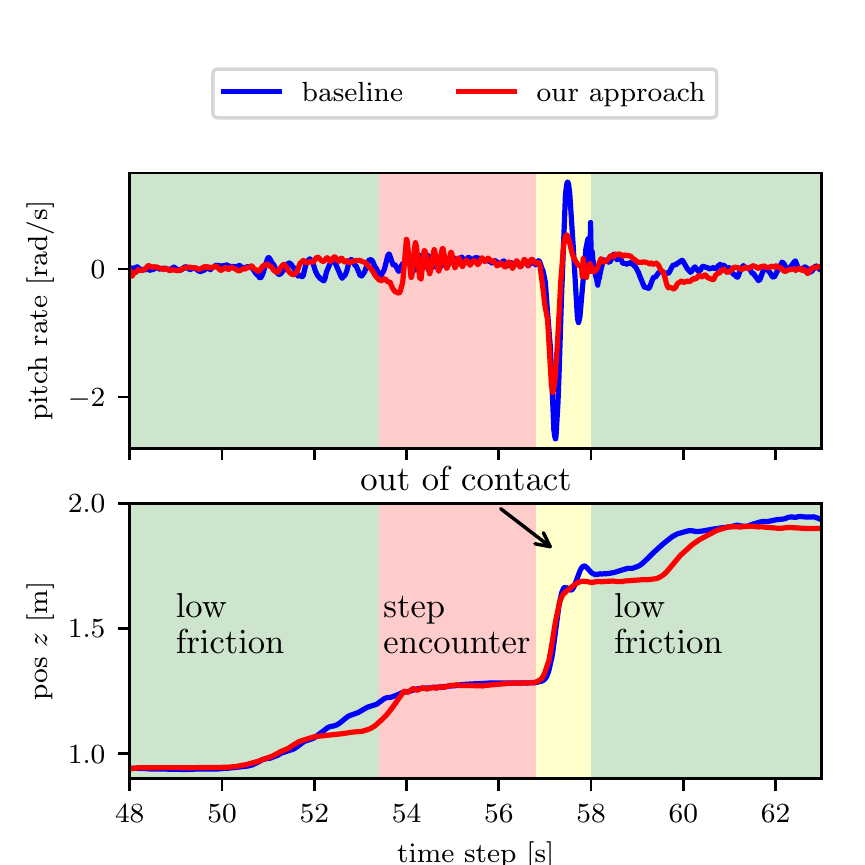}}
 \hfill
\subfloat{
\includegraphics[width=0.44\columnwidth,valign=c]{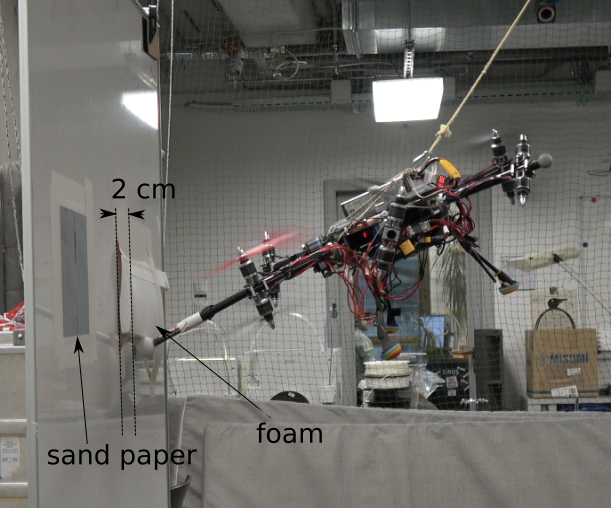}}
\hfill
\caption{The end effector is blocked by a step of 2 cm made out of foam while sliding
on a slippery whiteboard.}
\label{pic:Omav_sliding}
\end{figure}

The experiment (Fig.~\ref{pic:Omav_sliding}) presented in this section aims to investigate the ability to reject the disturbance caused by surface discontinuity and remain stable (criteria 3 in Sec.~\ref{subsec:problem_statement} ).
For the real experiment, a piece of foam is taped to the whiteboard (Fig.~\ref{pic:Omav_sliding}) and creates a step of about \SI{2}{\centi\metre} along the sliding trajectory.
During sliding, the step blocks the end-effector, which leads to an increase of the pitch angle. 
The end-effector then detachs from the surface and the sudden disappearance of the contact force presents as a large disturbance to the robot (see video attachment).

Our approach is trained as follows: Initially an even surface with randomly generated friction coefficients as described in Sec.~\ref{subsec:diffCoeff} is used to bootstrap the policy.
Then a piece of uneven surface with \SI{1}{\centi\metre} steps is created by setting each neighbouring surface to have a height difference of \SI{1}{\centi\metre}.
The policy is refined and trained for 900 epochs.
The height difference is then changed to \SI{2}{\centi\metre} with another 700 epochs of training.
Training was terminated early when the reward stopped increasing for 300 epochs.

The behaviour exhibited by our approach when a step is encountered is to adapt the control gain to be more compliant.
This leads to less oscillations in pitch velocity when the Omav's end effector slips and is out of contact (the yellow region in Fig.~\ref{pic:Omav_sliding}) compared to the baseline approach.



\reviewchanges{
\subsection{Sliding across a challenging surface}
A challenging rock-like papier-m\^ach\'e surface (Fig.~\ref{pic:Omav_sliding_rock}) is set up to compare our approach with baseline controllers. The surface of this rock is uneven and heterogeneous. 
Double-sided tapes and plastic surface with lubricant are added to the surface to emulate heterogeneous surface.
Two trajectories were tested for more variety.
For the training environment, we use procedurally generated terrain maps that have similar surface variations to the rock. 
During rollout, the fricition coefficients are randomly selected from six friction coefficients (0.1, 0.15, 0.45, 0.6, 0.75, 0.9) every four seconds.
Note that the evaluation and training environments are distinct and demonstrated the generalizability of our approach.}

\begin{figure}[t]
\centering
\includegraphics[width=0.8\columnwidth]{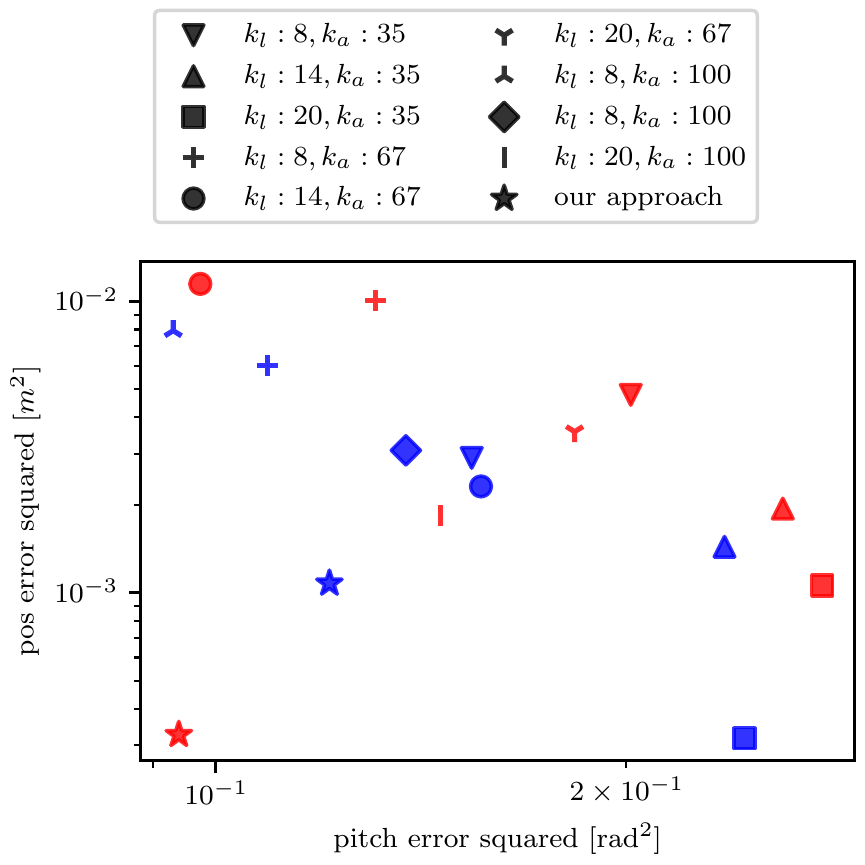}
\caption{\reviewchanges{Comparison of tracking performance of the learned policy with different combinations of constant gains. The learned policy outperforms all the constant gain performance. Note that it is plotted in logarithm scale for a better visualization of data. Two different colors are used to denote different trajectories. $k_a$ and $k_l$ denote the angular stiffness gain and translation stiffness gain, respectively. } }
\label{pic:rockSlideScatter}
\end{figure}

\reviewchanges{
As shown in Fig.~\ref{pic:rockSlideScatter}, our approach consistently outperforms baseline controller with nine combinations of low, middle and high angular and translation gain.
These different combinations are tested to see whether a tuning process that is often practiced in real-world can outperform our approach.
Each data point represents the average over the sliding of the same trajectory for the three times with the same controller.
Several interesting observations can be made: firstly, among the baseline approaches, a good tracking performance is generally achieved by high stiffness controller but the converse is not true. 
It is observed in the experiment that a high gain controller (both high in $k_l$ and $k_a$) can lead to instability.
However, we never experienced instability issues with our policy throughout the experiments.
This indicates that our approach adapts the controller to be more compliant when necessary.
Secondly, for each trajectory, the optimal set of constant gains are different. 
For example, to achieve best tracking performance in pitch, trajectory red and trajectory blue have different optimal gains (upper left in the plot).
This means that in reality, the engineer have to tune the parameters for each specific trajectory and surface to gain optimal performance. 
However, our approach always performs well for both trajectories. Our approach are both located in the lower left of the plot, which means good tracking performance in both position and orientation.
This implies that our policy adapts to the surface unevenness and varying friction coefficients.}

\begin{figure}[t]
\centering
\includegraphics[width=\columnwidth]{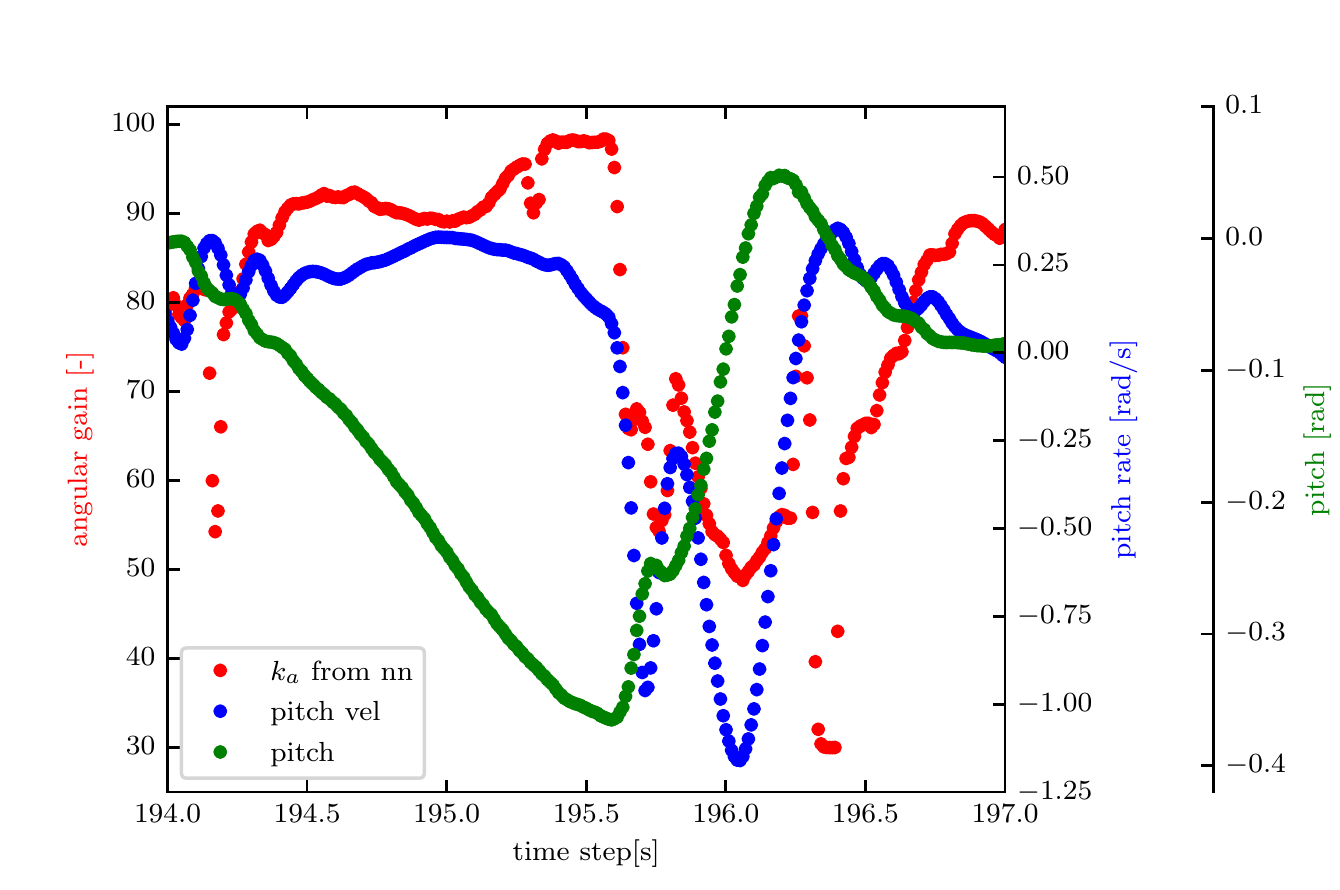}
\caption{\reviewchanges{An time history plot of partial inputs and outputs of the policy when sliding on the rock-like surface} }
\label{pic:rockSlideHistory}
\end{figure}

\reviewchanges{
An time history plot of partial inputs and outputs of the policy in plot Fig.~\ref{pic:rockSlideHistory} provides an intuition on how our policy behaves. Note that when the pitch error increases from 194 s to 195.5 s, the angular gain $k_a$ is also increased to maximum to reduce the tracking error. However, around 195.5, the end-effector detaches from the surface and causes oscillations on pitch rate, the angular gain is decreased to be more compliant.
}

\section{Conclusion}
This paper presented an approach to reject disturbances caused by discontinuous surface variations in geometry and friction during aerial sliding tasks for fully actuated flying vehicles.
When the environmental property changes, an adaptation policy adjusts the control gains of a standard impedance controller to reject these disturbances.
Experimental results demonstrated that the policy learned in simulation can be directly transferred to the aerial vehicle without adaptation. \reviewchanges{The learned policy is able to slide on a challenging rock-like surface and outperform state-of-art interaction controllers .}


\ifPrePrint
\else
	\thispagestyle{plain}
	\pagestyle{plain}
\fi

\IEEEpeerreviewmaketitle

\bibliographystyle{IEEEtran}
\bibliography{IEEEabrv,bibliography}
%

\end{document}

%% file: macros.tex
\newcommand{\reviewchanges}[1]{{#1}} 
\newcommand{\todo}[1]{\textcolor{red}{TODO: #1}}

\newcommand{\mvec}[1]{\boldsymbol{#1}}
\renewcommand{\vec}[1]{\bm{#1}} 

\newcommand{\actuatorCmd}{\mvec{u}}

\newcommand{\inertia}{\mvec{M}}
\newcommand{\damping}{\mvec{D}}
\newcommand{\stiffness}{\mvec{K}}

\newcommand{\feature}{\mvec{z}}

\newcommand{\zeroVector}{\bm{0}}
\newcommand{\oneVector}{\bm{1}}

\newcommand{\desInertia}{\inertia_\mathrm{des}}
\newcommand{\desDamping}{\damping_\mathrm{des}}
\newcommand{\desStiffness}{\stiffness_\mathrm{des}}
\newcommand{\desInertiaAdaptive}{\inertia_\mathrm{des}(\feature)}
\newcommand{\desDampingAdaptive}{\damping_\mathrm{des}(\feature)}
\newcommand{\desStiffnessAdaptive}{\stiffness_\mathrm{des}(\feature)}

\newcommand{\desStiffnessMin}{\stiffness_\mathrm{min}}
\newcommand{\desStiffnessMax}{\stiffness_\mathrm{max}}

\newcommand{\wrench}[1]{\mvec{w}_{#1}}

\newcommand{\dist}{\mathrm{dist}}
\newcommand{\ext}{\mathrm{dist}}
\newcommand{\contact}{\mathrm{con}}
\newcommand{\act}{\mathrm{act}}
\newcommand{\cmd}{\mathrm{cmd}}

\newcommand{\meas}{\mathrm{meas}}

\newcommand{\generalizedPos}{\mvec{q}}
\newcommand{\generalizedVel}{\tilde{\mvec{v}}}
\newcommand{\generalizedAcc}{\dot{\tilde{\mvec{v}}}}

\newcommand{\generalizedPosEst}{\hat{\mvec{q}}}
\newcommand{\generalizedVelEst}{\hat{\mvec{\dot{q}}}}

\newcommand{\generalizedPosRef}{\generalizedPos_\mathrm{ref}}
\newcommand{\generalizedVelRef}{\generalizedVel_\mathrm{ref}}
\newcommand{\generalizedAccRef}{\generalizedAcc_\mathrm{ref}}

\newcommand{\generalizedPosError}{\tilde{\mvec{e}}_s}
\newcommand{\generalizedVelError}{\tilde{\mvec{e}}_v}
\newcommand{\generalizedAccError}{\dot{\tilde{\mvec{e}}}_{v}}

\newcommand{\rot}[2]{\bm{R}^{{#1}{#2}}}

\newcommand{\base}{B}

\newcommand{\tool}{T}

\newcommand{\real}[2]{\mathbb{R}^{{#1}\times{#2}}}
\newcommand{\R}[1]{\mathbb{R}^{{#1}}}

\newcommand{\norm}[1]{\left\lVert#1\right\rVert}

\newcommand{\privilegedInfo}{\mvec{z}_\mathrm{gt}}

\newcommand{\optimalStudentPolicy}{\pi^{*}_\mathrm{s}}
\newcommand{\initialStudentPolicy}{\pi^*}
\newcommand{\teacherPolicy}{\pi_\mathrm{t}}
\newcommand{\observation}{\mvec{o}}

\newcommand{\eDist}{d}
\newcommand{\weight}{l}
\newcommand{\mdpState}{\mathcal{S}}
\newcommand{\mdpActionSpace}{\mathcal{A}}
\newcommand{\mdpRewardSpace}{\mathcal{R}}
\newcommand{\transitionProb}{P}
\newcommand{\action}{\vec{a}}
\newcommand{\reward}{r}
\newcommand{\priorMap}{\text{given surface map }}

%% file: images/learn_from_simulation.tex
\begin{tikzpicture}
		\draw [black] (0,0) rectangle (5,0.5);
		\node at (2.5,0.25){Design of teacher};
		\draw[->, thick] (2.5,0) -- (2.5, -0.5);
		\node at (3.25, - 0.25){Teacher};
		\draw [black] (0,-1.0) rectangle (5,-0.5);
		\node at (2.5,-0.75){Student teacher learning};
		\draw[->, thick] (2.5,-1.0) -- (2.5, -1.75);
		\node at (3.25, - 1.25){Policy};
		\draw [dashed] (-1.5,-1.5) -- (6.5,-1.5);
		\node at (-1, - 1.25){Simulation};
		\draw [black] (0,-2.25) rectangle (5,-1.75);
		\node at (-1, - 1.75){Real-world};

		\node at (2.5,-2.0){Deployment on real robot};
\end{tikzpicture}